%% file: main.tex
\pdfoutput=1

\documentclass[11pt]{article}

\usepackage{EMNLP2022}

\usepackage{times}
\usepackage{latexsym}

\usepackage[utf8]{inputenc} 
\usepackage[T1]{fontenc}    
\usepackage{hyperref}       
\usepackage{url}            
\usepackage{booktabs}       
\usepackage{nicefrac}       
\usepackage{microtype}      
\usepackage{wrapfig}

\usepackage{amsmath,amssymb,amsfonts}
\usepackage{algorithmic}
\usepackage{graphicx}
\usepackage{textcomp}
\usepackage{pifont}
\newcommand{\crossmark}{\ding{53}}%
\newcommand{\heavycheckmark}{\ding{52}}%
\usepackage{multirow}
\usepackage{xspace}
\newcommand{\minus}{\scalebox{0.75}[1.0]{$-$}}
\def\BibTeX{{\rm B\kern-.05em{\sc i\kern-.025em b}\kern-.08em
    T\kern-.1667em\lower.7ex\hbox{E}\kern-.125emX}}

\newcommand{\tekgen}{\textsc{TekGen}\xspace} 

\usepackage{subcaption}
\usepackage{url}

\title{Knowledge Graph Generation From Text}

\author{Igor Melnyk, Pierre Dognin, Payel Das\\
IBM Research \\
Yorktown Heights, NY 10598, USA \\
\texttt{igor.melnyk@ibm.com, pdognin@us.ibm.com, daspa@us.ibm.com}
}

\begin{document}
\maketitle
\begin{abstract}
In this work we propose a novel end-to-end multi-stage Knowledge Graph (KG) generation system from textual inputs, separating the overall process into two stages. The graph nodes are generated first using pretrained language model, followed by a simple edge construction head, enabling efficient KG extraction from the text. For each stage we consider several architectural choices that can be used depending on the available training resources. We evaluated the model on a recent WebNLG 2020 Challenge dataset, matching the state-of-the-art performance on text-to-RDF generation task, as well as on New York Times (NYT) and a large-scale \tekgen datasets, showing strong overall performance, outperforming the existing baselines. We believe that the proposed system can serve as a viable KG construction alternative to the existing linearization or sampling-based graph generation approaches.  Our code can be found at \url{https://github.com/IBM/Grapher}
\end{abstract}

\section{Introduction}
\label{sec:introduction}
\input{introduction}

\section{Method}
\label{sec:method}
\input{Method}

\section{Data}
\label{sec:data}
\input{data}

\section{Experiments}
\label{sec:experiments}
\input{experiments}

\section{Conclusion}
In this work, we proposed Grapher, a novel multi-stage KG generation system, that separates the overall graph generation into two steps. In the first step, the nodes are generated from the input text using a pretrained language model. The resulting node features  are then used for edge generation to construct the output graph. We proposed several architectural choices for each stage. In particular, graph nodes can either be generated as a sequence of text tokens or as a set of query-based feature vectors decoded into tokens through generation head (e.g., GRU). Edges can be either generated by a GRU decoding head or selected by a classification head. We also addressed the problem of skewed edge distribution, where the token/class corresponding to the missing edge is over-represented, leading to inefficient training. For this, we proposed to use of  either the focal loss, or the sparse adjacency matrix. The experimental evaluations showed that Grapher matched state-of-the-art performance on smaller WebNLG dataset, and showed strong overall performance, outperforming existing baselines, on NYT and \tekgen datasets, serving as a viable alternative to the existing baselines. 

\section*{Limitations}
There are several limitations of this work that need to be addressed in the future work. The first is the computational complexity of edge generation, which is quadratic in the number of edges, and this sets the limit on the sizes of the graphs that the systems can process. Moreover, since the nodes are generated using transformer-based models, which have quadratic complexity of the attention mechanism, there is a limit on the size of the input text the system can handle. Therefore, the current algorithm is suitable for small or medium size graphs and text passages. The extension to large scale is important and will be a part of the future effort. Moreover, the current setup was applied only to English domain datasets, which is a limitation, given that there is a benefit of multi- and cross-lingual training of language systems as ours. Finally, although not being our objective, the current model is designed to handle only the direction from text to  knowledge graph, and the reverse direction has not been explored yet but can be a part of the future investigation.

\bibliography{references}
\bibliographystyle{acl_natbib}

\clearpage

\appendix

\section{Appendix}

In Tables \ref{tab:meanstd_webnlg} and \ref{tab:meanstd_nyt} we present the results of the best performing Grapher configurations, which uses Text Nodes with either Class Edges or Gen Edges respectively, with multiple random initializations to examine the results variability on WebNLG and NYT test set. As can be seen, the scores averaged across 5 runs (with different random initializations) show low standard deviation, further validating Grapher's stable performance. 

\begin{table}[h]
\centering
\caption{Mean and standard deviation for the results of 5 randomly initialized runs of the best Grapher configuration which uses Text Nodes and Class Edges on WebNLG test set.}
\label{tab:meanstd_webnlg}
\bgroup
\def\arraystretch{1.0}%
\begin{tabular}{cccc}
  \multicolumn{1}{c}{\textbf{Match}} &
  \multicolumn{1}{c}{\textbf{F1}} &
  \multicolumn{1}{c}{\textbf{Precision}} &
  \multicolumn{1}{c}{\textbf{Recall}} \\ \hline
  \multicolumn{1}{l|}{Exact}       & $0.720_{\pm{0.05}}$  & $0.711_{\pm{0.05}}$ & $0.729_{\pm{0.06}}$ \\
  \multicolumn{1}{l|}{Partial}     & $0.744_{\pm{0.04}}$  & $0.737_{\pm{0.03}}$ & $0.763_{\pm{0.03}}$ \\
  \multicolumn{1}{l|}{Strict}      &  $0.716_{\pm{0.05}}$ & $0.709_{\pm{0.05}}$ & $0.724_{\pm{0.05}}$ \\ \hline
\end{tabular}
\egroup
\end{table}

\begin{table}[h]
\centering
\caption{Mean and standard deviation for the results of 5 randomly initialized runs of the best Grapher configuration which uses Text Nodes and Gen Edges on NYT test set.}
\label{tab:meanstd_nyt}
\bgroup
\def\arraystretch{1.0}%
\begin{tabular}{cccc}
  \multicolumn{1}{c}{\textbf{Match}} &
  \multicolumn{1}{c}{\textbf{F1}} &
  \multicolumn{1}{c}{\textbf{Precision}} &
  \multicolumn{1}{c}{\textbf{Recall}} \\ \hline
  \multicolumn{1}{l|}{Exact}       & $0.908_{\pm{0.06}}$  & $0.909_{\pm{0.05}}$ & $0.910_{\pm{0.04}}$ \\
  \multicolumn{1}{l|}{Partial}     & $0.910_{\pm{0.03}}$  & $0.910_{\pm{0.04}}$ & $0.908_{\pm{0.05}}$ \\
  \multicolumn{1}{l|}{Strict}      &  $0.905_{\pm{0.04}}$ & $0.903_{\pm{0.04}}$ & $0.904_{\pm{0.05}}$ \\ \hline
\end{tabular}
\egroup
\end{table}

In Tables \ref{tab:webnlg_extra} and \ref{tab:nyt_extra} we also present additional experiments by varying the T5 model size. In particular, in addition to the T5-large model, containing 770M parameters, used in the main paper, we also considered T5-base (220M parameters) and T5-small (60M parameters). It can be seen, that in general the performance drops as the model size decreases. However, for NYT dataset, the model architecture that uses Text Nodes and Class Edges, T5-small actually outperforms T5-base. At the same time, for Gen Edges all three model choices performed very similar with minor drop in performance as the size decreases.

\begin{table}[h]
\centering
\caption{WebNLG}
\label{tab:webnlg_extra}
\resizebox{0.48\textwidth}{!}{
\begin{tabular}{cccccc}
 &
   &
   &
  \textbf{Large}&
  \textbf{Base} &
  \textbf{Small} \\\cline{3-6}
\multirow{3}{*}{\rotatebox{90}{\textbf{Grapher}}} &
  \multirow{3}{*}{\begin{tabular}[c]{@{}c@{}}Text\\ Nodes\end{tabular}} &
  \multicolumn{1}{c|}{\begin{tabular}[c]{@{}c@{}}Gen\\ Edges\end{tabular}} &
  \textbf{0.683} &
  0.660 &
  0.596 \\ \cline{3-6} 
 &
   &
  \multicolumn{1}{c|}{\begin{tabular}[c]{@{}c@{}}Class\\ Edges\end{tabular}} &
  \textbf{0.722} &
  0.693 &
  0.631
\end{tabular}
}
\end{table}

\begin{table}[h]
\centering
\caption{NYT}
\label{tab:nyt_extra}
\resizebox{0.48\textwidth}{!}{
\begin{tabular}{cccccc}
 &
   &
   &
  \textbf{Large}&
  \textbf{Base} &
  \textbf{Small} \\\cline{3-6}
\multirow{3}{*}{\rotatebox{90}{\textbf{Grapher}}} &
  \multirow{3}{*}{\begin{tabular}[c]{@{}c@{}}Text\\ Nodes\end{tabular}} &
  \multicolumn{1}{c|}{\begin{tabular}[c]{@{}c@{}}Gen\\ Edges\end{tabular}} &
  \textbf{0.912} &
  0.907 &
  0.897 \\ \cline{3-6} 
 &
   &
  \multicolumn{1}{c|}{\begin{tabular}[c]{@{}c@{}}Class\\ Edges\end{tabular}} &
  \textbf{0.870} &
  0.812 &
  0.846
\end{tabular}
}
\end{table}

\begin{figure*}[h!]
\centering
\includegraphics[width=0.8\textwidth]{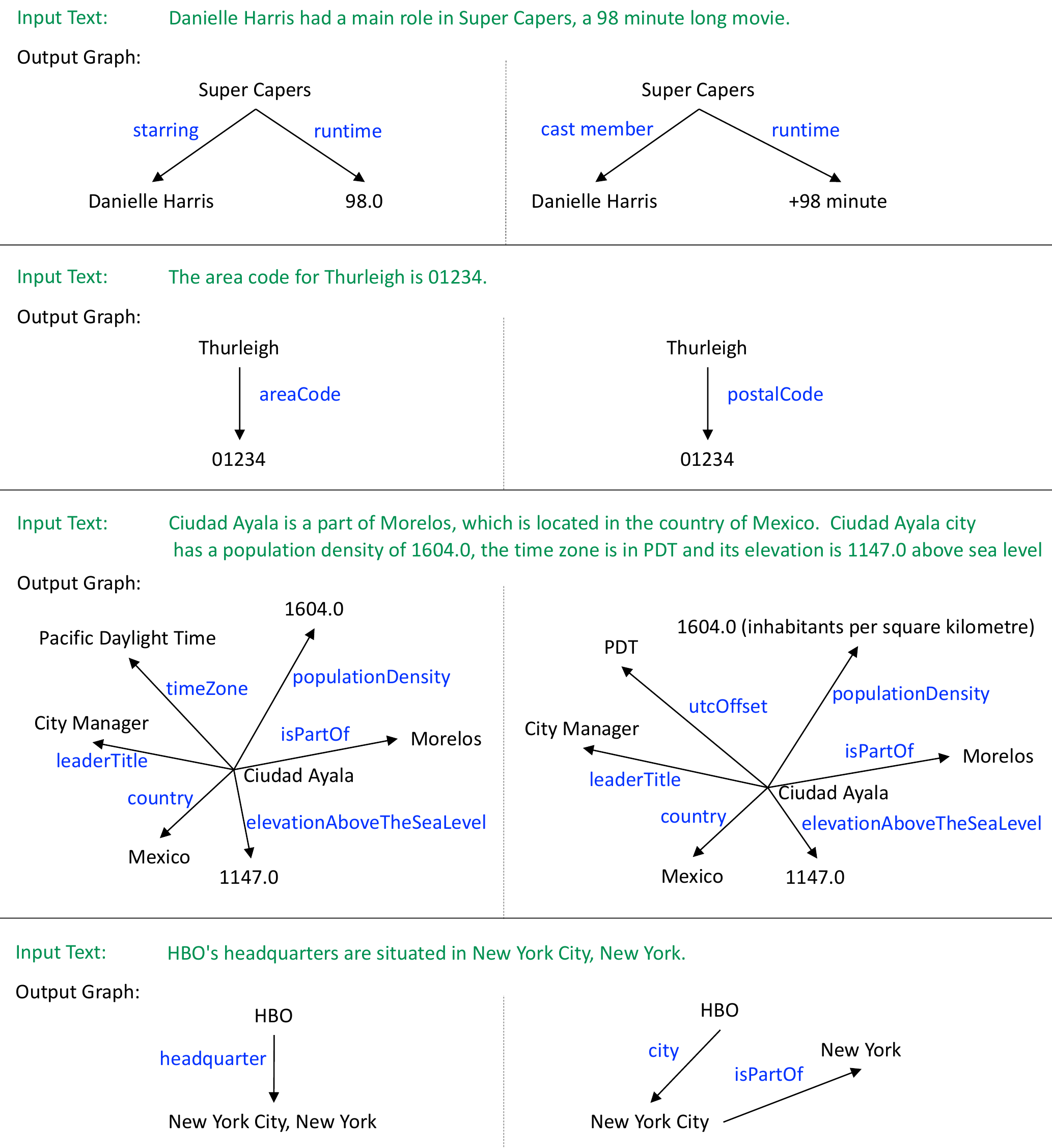}
\caption{Examples of some of the notable generated (right column) and the ground truth graphs (left column) for WebNLG dataset.}
\label{fig:ex_webnlg}
\end{figure*}

\begin{figure*}[h!]
\centering
\includegraphics[width=0.8\textwidth]{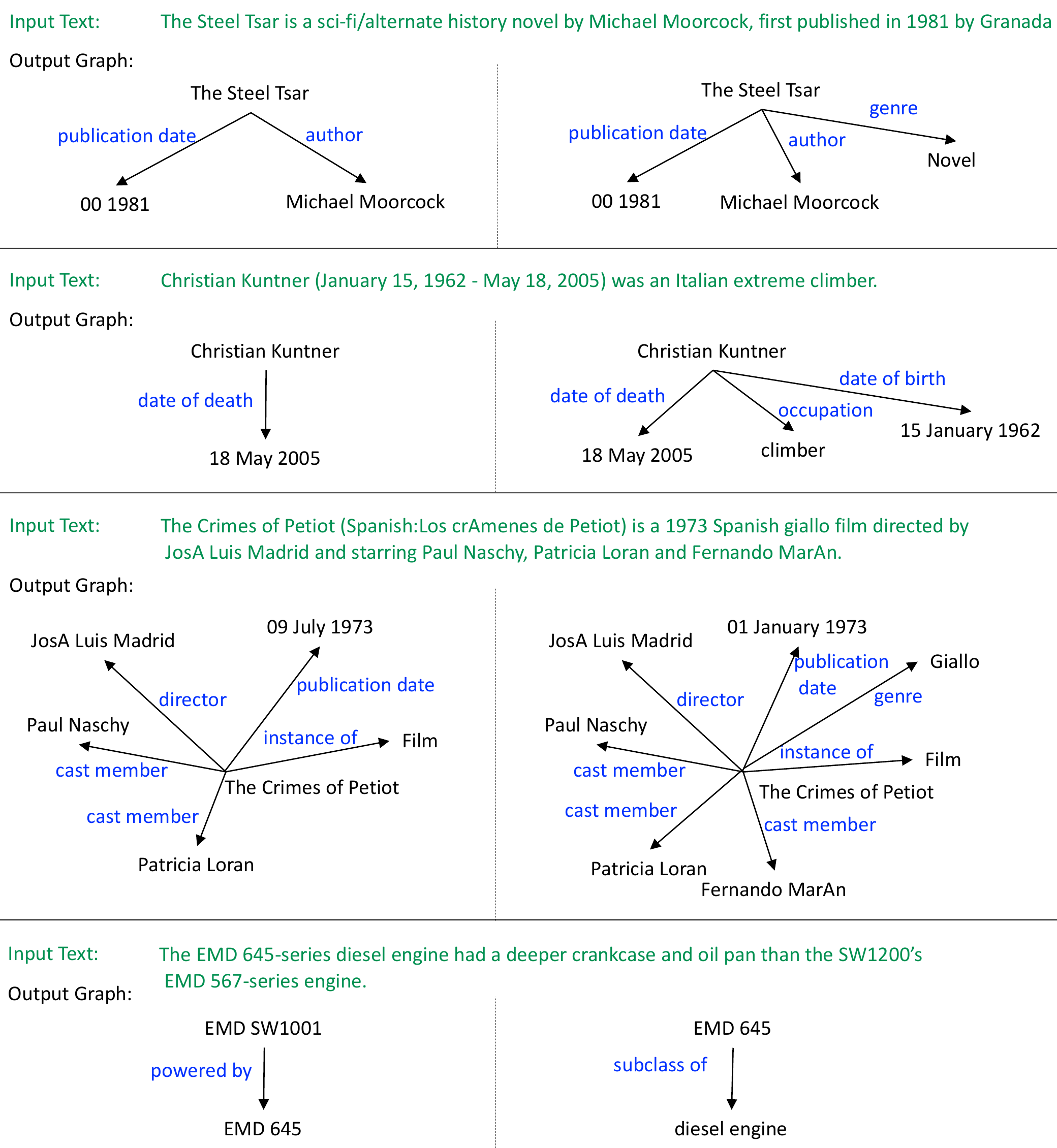}
\caption{Examples of some of the notable generated (right column) and the ground truth graphs (left column) for \tekgen dataset.}
\label{fig:ex_tekgen}
\end{figure*}

In Fig.~\ref{fig:ex_webnlg} we present some examples of the generated graphs (right column) and their associated ground truths (left column) for WebNLG dataset. In Fig.~\ref{fig:ex_tekgen} similar results are given for \tekgen dataset. Both examples show that the trained Grapher system sometimes can generated more detailed and accurate graphs corresponding to the input text as compared to the ground truth (e.g., first three examples in Fig.~\ref{fig:ex_tekgen}, where it adds extra edges for genre, occupation and birth date). Also, the use of T5 model for node extraction shows that the model can include information in the generated nodes that is not present in the input text (e.g., third example in Fig.~\ref{fig:ex_webnlg}, which included `inhabitants per square kilometre', possibly from T5's original pre-training on large textual corpora.)

\end{document}

%% file: introduction.tex
Automatic Knowledge Graph (KG) construction is an active research area aiming at representing the information present in abundant textual corpora in a more organized, structured and compressed form, which can be efficiently utilized in a variety of downstream applications, including reasoning, decision making, question answering, to name a few. However, this is a challenging problem due to the inherent non-unique graph representation (graph with $N$ nodes can have $N!$ equivalent adjacency matrices), complex node and edge structure (node set is not fixed and edges are not binary), large output spaces (for graph with $N$ nodes the system may need to output up to $N^2$ edges to specify its structure), lack of efficient architectures specialized for graph-structured generation output and limited parallel training data.

The related problem of generating text from a given KG is generally more widely studied, with many suggested architectures and approaches. Among the proposed methods, some of the current state-of-the-art systems that work on small or moderately-sized graphs, \cite{Li2020LeveragingLP, Ribeiro2020InvestigatingPL, agarwal-etal-2020-machine, UnifiedSKG}, usually formulate it as a simple sequence-to-sequence problem by representing the graph in a linearized form and fine-tune the pre-trained language models (PLMs), such as T5 \cite{raffel2020exploring} or BART \cite{Lewis2020BARTDS}, on the task of translating the sequence of triples to the corresponding textual description.

\begin{figure}[t!]
\centering
\includegraphics[width=0.48\textwidth]{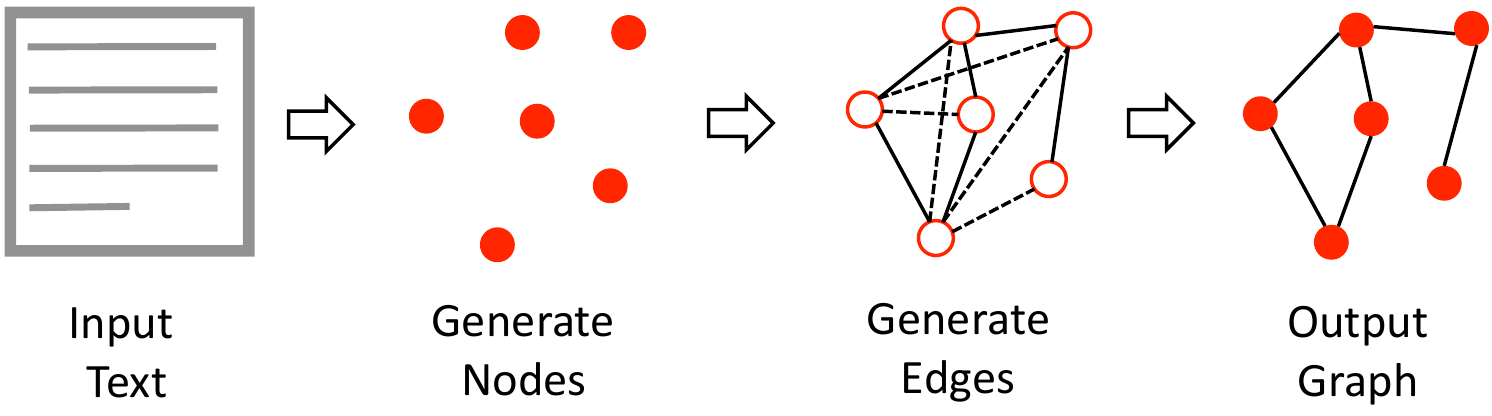}
\caption{Grapher overview. For a given text input, in the first step we generate graph nodes, leveraging the representation power of pre-trained language models, fine-tuned on the task of entity extraction. In the second step, the graph edges are generated using the available entity information to construct the final graph.}
\label{fig:overview}
\end{figure} 

Nevertheless, KG generation remains a popular research area, receiving attention from many communities, including natural language processing (NLP), data mining, and machine learning. Recent success of the Transformer-based language models from the NLP community \cite{vaswani2017attention, Devlin2019BERTPO, raffel2020exploring},  pre-trained on large textual corpora, led to a series of works that attempted to exploit the vast amounts of learned linguistic knowledge for the downstream task of KG construction. Some of these approaches looked into a simpler problem of graph completion \cite{conceptnet-600k, yao2019kgbert, malaviya2020commonsense}.
The drawback of these methods is that they are limited to the task of extending existing graphs by local neighborhood modifications and are not suitable for building the entire global graph structures. Alternatively, other works \cite{Petroni2019LanguageMA, Roberts2020HowMK, Jiang2019HowCW, Shin2020AutoPromptEK, Li2021PrefixTuningOC} proposed to query the pre-trained models to extract the learned factual and commonsense knowledge. The idea is to prompt the language model to predict the masked objects in cloze sentences describing the partially complete triples. Similarly as before, these methods are usually only suitable for local graph patching, lacking the ability to perceive the global graph structure.


Alternatively, there are a number of works that propose to generate the entire graph structure ground up. One example is GraphRNN from \citet{You2018GraphRNNGR}, which models a graph as a sequence of additions of new nodes using node-level RNN and edges using another edge-level RNN. Although promising for our task of KG construction, the sequential and greedy nature of its generation can cause sub-optimal graph structures. CycleGT of \cite{Guo2020CycleGTUG} is an unsupervised method for text-to-graph and graph-to-text generation, where the graph generation part relies on off-the-shelf entity extractor followed by a classifier to predict the relationships. The reliance on external NLP pipelines breaks the end-to-end continuity of system training, potentially leading to sub-optimal results. Similarly, \cite{dognin-etal-2020-dualtkb} proposed DualTKB employing unsupervised cycle loss to enable the graph-text translation in both directions. However, their method was applied only to single sentence-single triple generation, limiting applicability for larger graphs. Other approaches, such as BT5 from \cite{agarwal-etal-2020-machine} proposed to utilize large pre-trained T5 model to generate KG in a linearized form, where the object-predicate-subject triples are concatenated together and the entire text-to-graph problem is viewed as sequence-to-sequence modeling. The potential issue with this approach is that the graph linearization is not unique and inefficient due to the repetition of graph components multiple times, leading to long sequences and increased complexity. \cite{lu-etal-2022-unified} is another text-to-structure method, however it uses predefined schema (e.g., for entity or triplet extraction), while our method is schema-free and generalizes to any text form of nodes and edges. Finally, \cite{Wang2020LanguageMA} proposed MaMa for KG construction, where entities and relationships are first matched using the attention weight matrices from the forward pass of the LM. Those are then mapped to the existing KG schema to generate the final graph.


\textbf{The proposed system: Grapher} Analyzing the shortcomings of the existing methods, in this work we propose to address them with a novel Knowledge Graph construction system which we call Grapher, presented schematically in Fig.~\ref{fig:overview}.  Given input text, the graph generation is split into two steps. In the first step, we leverage the representation power of pre-trained language models, e.g., T5 \cite{raffel2020exploring}, fine-tuned on the task of entity (graph nodes) extraction, while in the second stage the relationships (graph edges) are generated using the available entity information. There are three main properties of Grapher: \textbf{(i)} The use of state-of-the-art language models pre-trained on large textual corpora, used for node generation is key to the algorithm's performance as it lays out the foundation for the entire graph. The available parallel data for learning the text to graph translation is usually small, therefore training custom-built entity extraction architectures from scratch on this limited data is inferior to fine-tuning the already pretrained Transformer-based language models. \textbf{(ii)} The partitioning of graph construction process into two steps ensures efficiency that each node and edge is generated only once, which is in contrast to graph linearization approaches, e.g.,  \cite{agarwal-etal-2020-machine} \cite{dognin2021regen}, whose graph sequence representation is non-unique and can be inefficient. \textbf{(iii)} Finally, the entire system is end-to-end trainable, where the node and edge generation are optimized jointly, enabling efficient information transfer between the two modules, avoiding the need of any external NLP pipelines such as entity/relation extraction, co-reference resolution, etc.  We evaluate the proposed Grapher on three datasets: the WebNLG+ 2020 Challenge \cite{Ferreira2020The2B} matching state-of-the-art performance for Text-to-RDF generation as well as on NYT \cite{riedel2010modeling} and a recent large-scale \tekgen \cite{agarwal2021knowledge} dataset showing strong results outperforming existing baselines.


%% file: method.tex
\begin{figure}[th!]
\centering
\includegraphics[width=0.45\textwidth]{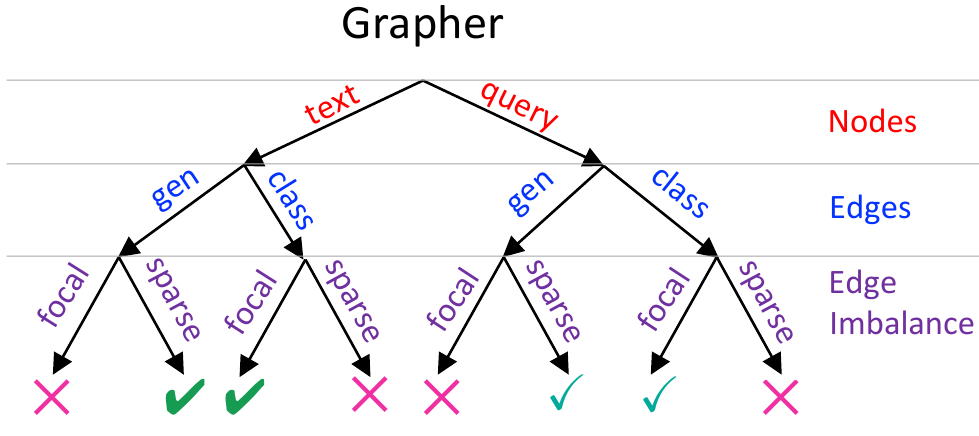}
\caption{Grapher architectural choices. \textcolor{red}{\crossmark} - setups that did not show advantage or did not perform well during preliminary evaluations,  \textcolor{green}{\checkmark} - selected for further evaluation , \textcolor{green}{\heavycheckmark} - best performing system}
\label{fig:summary}
\end{figure}

In this Section we cover the details of the proposed approach, first describing the functionality of the node generation in Section \ref{sec:node_gen}, followed by the edge generation in Section~\ref{sec:edge_gen} and the discussion on edge imbalance problem in Section~\ref{sec:edge_imb}.  In Fig.~\ref{fig:summary} we summarize all the architectural choices of the Grapher system. The branches marked with a red cross denote the setups which in our earlier evaluations did not show advantage over the neighboring branch, e.g., the focal loss underperformed the sparse edge training for the text nodes combined with edge generation head. The branches with green check marks are the ones we select for further evaluation. The bold dark green check marks show two best performing systems across multiple experiments. In what follows, we now show the details of these choices.

\subsection{Node Generation: Text Nodes}
\label{sec:node_gen}

Given text input, the objective of this module is to generate a set of unique nodes, which define the foundation of the graph. As we mentioned in Section \ref{sec:introduction}, the node generation is key to the successful operation of Grapher, therefore for this task we use a pre-trained encoder-decoder language model (PLM), such as T5. Using a PLM, we can now formulate the node generation as a sequence-to-sequence problem, where the system is fine-tuned to translate textual input to a sequence of nodes, separated with special tokens, 
$\langle \textsc{pad} \rangle ~\textsc{node}_1~ \langle \textsc{node\_sep}\rangle ~\textsc{node}_2~ \cdots \langle /\textsc{s}\rangle$, where $\textsc{node}_i$ represents one or more words.

\begin{figure}[th!]
\begin{minipage}[b]{0.48\textwidth}
\centering
\includegraphics[width=0.85\textwidth]{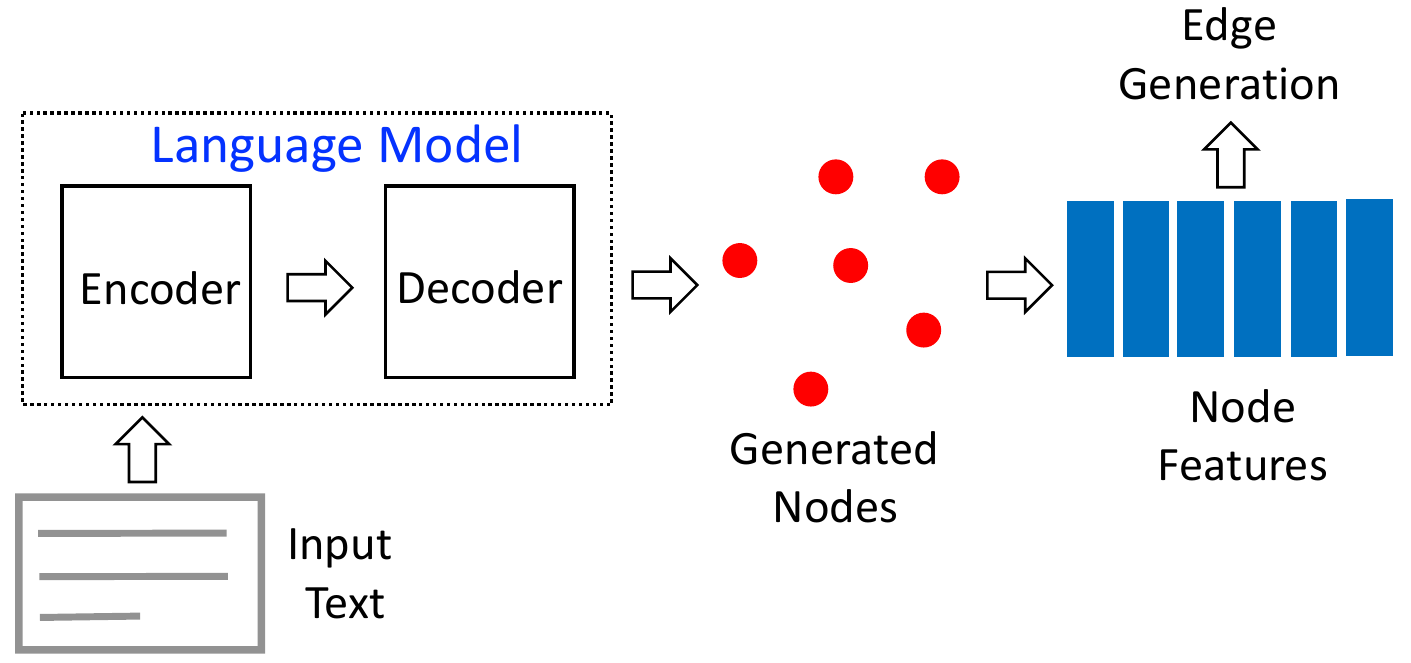}
\caption{Node generation using traditional sequence-to-sequence paradigm based on T5 language model, where the input text is transformed into a sequence of text entities. The features corresponding to each entity (node) is extracted and sent to the edge generation module.}
\label{fig:text_nodes}
\end{minipage}
\hspace{0.5em}
\begin{minipage}[b]{0.48\textwidth}
\centering
\includegraphics[width=0.99\textwidth]{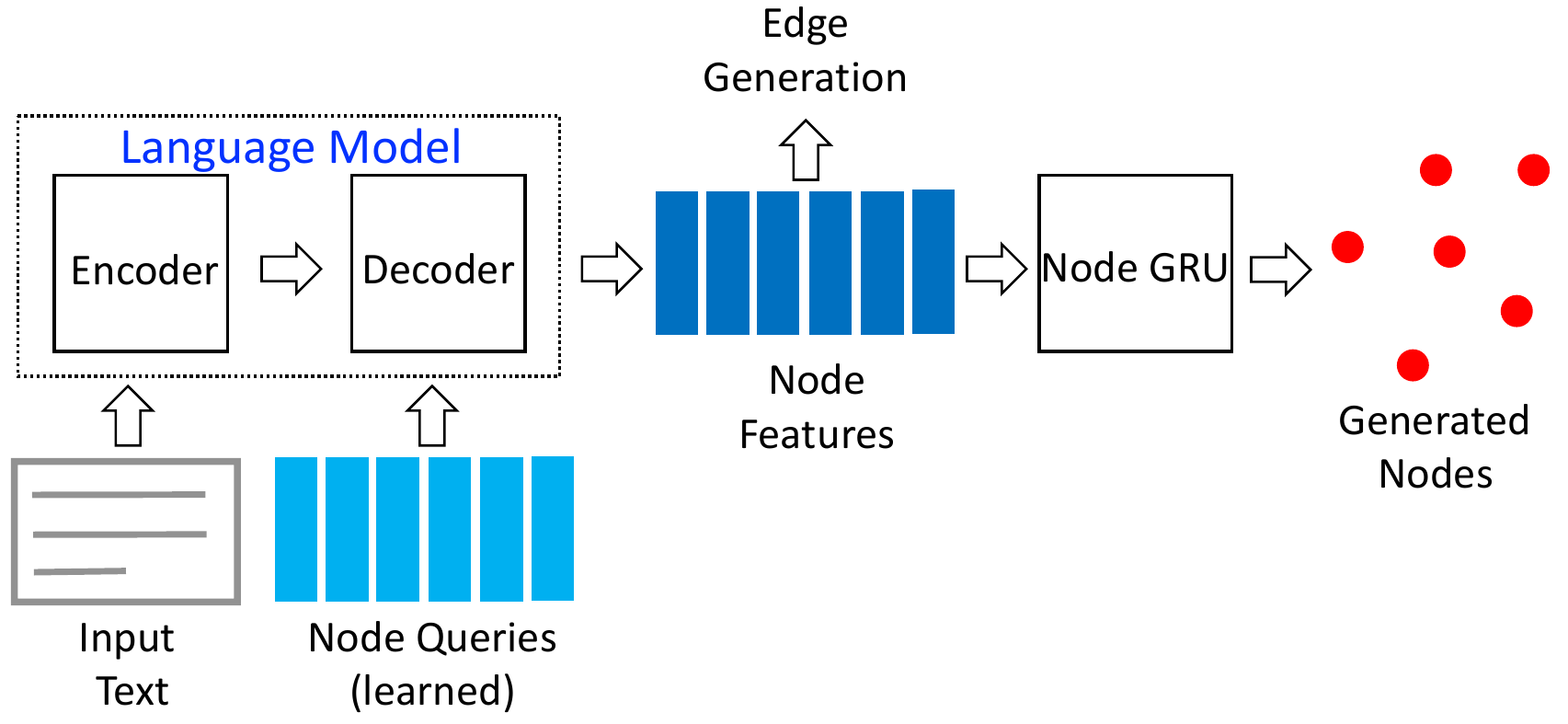}
\caption{Node generation using learned query vectors. Here the input text and the query vectors (in the form of embedding matrix) is transformed into node features. Those are then decoded into graph nodes using node generation head (e.g, LSTM or GRU). The same features are also sent to the edge construction module.}
\label{fig:query_nodes}
\end{minipage}
\end{figure}

As seen in Fig.~\ref{fig:text_nodes}, in addition to node generation, this module supplies node features for the downstream task of edge generation. Since each node can have multiple associated words, we greedy-decode the generated string and utilize the separation tokens $\langle \textsc{node\_sep}\rangle$ to delineate the node boundaries and mean-pool the hidden states of the decoder's last layer. Note that in practice we fix upfront the number of generated nodes and fill the missing ones with a special $\langle \textsc{no\_node}\rangle$ token.

\subsection{Node Generation: Query Nodes}
\label{sec:query_nodes}

One issue with the above approach is ignoring that the graph nodes are permutation invariant, since any permutation of the given set of nodes should be treated equivalently. To address this limitation, we propose a second architecture, inspired by DETR \cite{Carion2020EndtoEndOD}. See Fig.~\ref{fig:query_nodes} for an illustration.

\textit{Learnable Node Queries} The decoder receives as input a set of learnable node queries, represented as an embedding matrix. We also disable causal masking, to ensure that the Transformer is able to attend to all the queries simultaneously. This is in contrast to the traditional encoder-decoder architecture that usually gets as an input embedding of the target sequence with the causal masking during training or the embedding of the self-generated sequence during inference. The output of the decoder can now be directly read-off as $N$ (number of nodes) $d$-dimensional node features $F_n \in \mathbb{R}^{d\times N}$ and passed to a prediction head (LSTM or GRU) to be decoded into node logits $L_n \in \mathbb{R}^{S\times V\times N}$, where $S$ is the generated node sequence length and $V$ is the vocabulary size. 

\textit{Permutation Matrix} To avoid the system to memorize the particular target node order and enable permutation-invariance, the logits and features are permuted as
\begin{align}
    L_n^\prime(s) = L_n(s)P,  ~~F_n^\prime = F_nP,
\end{align}
for $s = 1,\ldots,S$ and where $P \in \mathbb{R}^{N \times N}$ is a permutation matrix obtained using bipartite matching algorithm between the target and the greedy-decoded nodes. We used cross-entropy loss as the matching cost function. The permuted node features $F_n^\prime$ are now target-aligned and can be used in the edge generation stage. 

\subsection{Edge Generation}
\label{sec:edge_gen}

\begin{figure}[th!]
\begin{minipage}[b]{0.48\textwidth}
\centering
\includegraphics[width=0.99\textwidth]{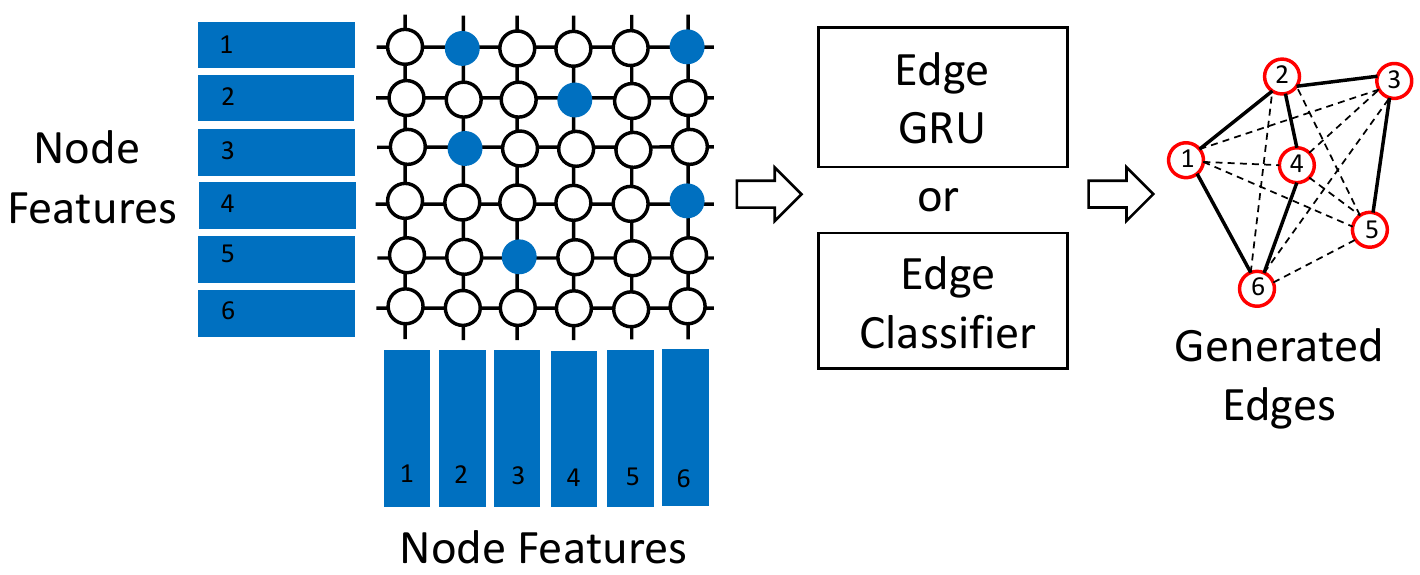}
\caption{Edge construction, using generation (e.g., GRU) or a classifier head. Blue circles represent the features corresponding to the actual graph edges (solid lines) and the white circles are the features that are decoded into $\langle \textsc{no\_edge}\rangle$ (dashed line).}
\label{fig:edges_full}
\end{minipage}
\hspace{0.5em}
\begin{minipage}[b]{0.48\textwidth}
\centering
\includegraphics[width=0.99\textwidth]{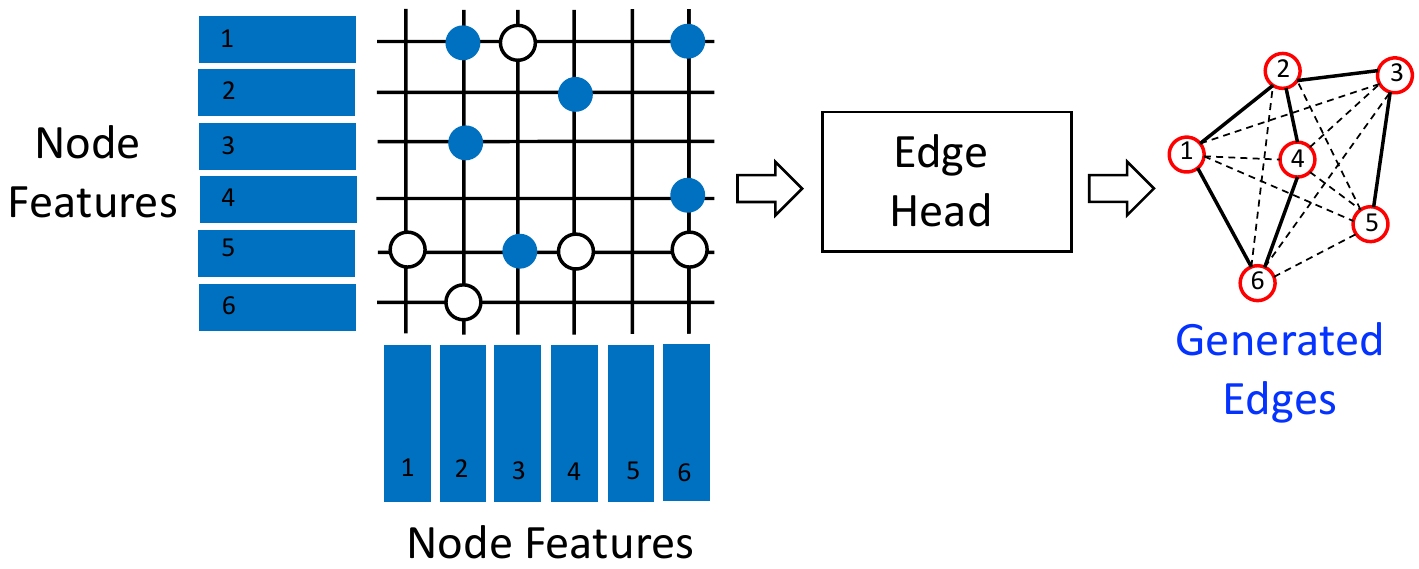}
\caption{Edge generation with sparse adjacency matrix, using same decoder heads as in Fig.~\ref{fig:edges_full}. Here while keeping all the actual edges, we remove most of the $\langle \textsc{no\_edge}\rangle$ tokens, leaving only a few. This setup is only used during training to improve the edge imbalance problem and speedup the training.}
\label{fig:edges_sparse}
\end{minipage}
\end{figure}

The generated set of node features from previous step is then used in this module for the edge generation. Fig.~\ref{fig:edges_full} shows a schematic description of this step. 
Given a pair of node features, a prediction head decides the existence (or not) of an edge between their respective nodes.
One option is to use a head similar to the one in Section \ref{sec:query_nodes} (LSTM or GRU) to generate edges as a sequence of tokens. The other option is to use a classification head to predict the edges. The two choices have their own pros and cons and the selection depends on the application domain. The advantage of generation is the ability to construct \emph{any} edge sequence, including ones unseen during training, at the risk of not matching the target edge token sequence exactly. On the other hand, if the set of possible relationships is fixed and known, the classification head is more efficient and accurate, however if the training has limited coverage of all possible edges, the system can misclassify during inference. We explore both options in Section \ref{sec:experiments}.

Note that since in general KGs are represented as directed graphs, it is important to ensure the correct order (subject-object) between two nodes. For this, we propose to use a simple difference between the feature vectors: $F_n^{\prime}(:,i)- F_n^{\prime}(:,j)$ for the case when the node $i$ is a parent of node $j$. We experimented with other options, including concatenation and adding position information but found the difference being the most effective, since the model learns that $F_n^{\prime}(:,i)- F_n^{\prime}(:,j)$ implies $i\rightarrow j$, while $F_n^{\prime}(:,j)- F_n^{\prime}(:,i)$ implies $j\rightarrow i$.

\subsection{Imbalanced Edge Distribution}
\label{sec:edge_imb}

Observe that since we need to check the presence of edges between all pairs of nodes, we have to generate or predict up to $N^2$ edges, where $N$ is the number of nodes. There are small savings that can be done by ignoring self-edges as well as ignoring edges when one of the generated nodes is the $\langle \textsc{no\_node}\rangle$ token. When no edge is present between the two nodes, we denote this with a special token $\langle \textsc{no\_edge}\rangle$.  Moreover, since in general the number of actual edges is small and $\langle \textsc{no\_edge}\rangle$ is large, the generation and classification task is imbalanced towards the $\langle \textsc{no\_edge}\rangle$ token/class. To remedy this, we propose two solutions: one is a modification of the cross-entropy loss, and the other is a change in the training paradigm.

\textbf{Focal Loss} Here we replace the traditional Cross-Entropy (CE) loss with Focal (F) loss \cite{Lin2020FocalLF}, whose main idea is down-weight the CE loss for well-classified samples ($\langle \textsc{no\_edge}\rangle$) and increase the CE loss for mis-classified ones, as illustrated below for a  probability $p$ corresponding to a single edge and $t$ is a target class: 
\begin{align*}
    \text{CE} = \minus\log(p_t), ~\text{F} = \minus(1\minus p_t)^\gamma\log(p_t),
\label{eq:focal}    
\end{align*}
where $\gamma \geq 0$ is a weighting factor, such that $\gamma = 0$ makes both losses equivalent. The application of this loss to the classification head is straightforward while for the generation head we modify it by first accumulating predicted probabilities over the edge sequence length to get the equivalent of $p_t$ and then apply the loss. In practice, we observed that Focal loss improved the accuracy for the classification head, while for the generation head the performance did not change significantly.


\textbf{Sparse Edges} To address the edge imbalance problem another solution is to modify the training settings by sparsifying the adjacency matrix to remove most of the $\langle \textsc{no\_edge}\rangle$ edges as shown in Fig.~\ref{fig:edges_sparse}, therefore re-balancing the classes artificially. Here, we keep all the actual edges but then leave only a few randomly selected $\langle \textsc{no\_edge}\rangle$ ones. Note that this modification is done only to improve efficiency of the training, during inference the system still needs to output all the edges, as in Fig.~\ref{fig:edges_full}, since their true location is unknown. In practice, besides seeing 10-20\% improvement in accuracy, we also observed about 10\% faster training time when using sparse edges as compared to using full adjacency matrix.

%% file: data.tex
To evaluate Grapher's performance and compare it to the baselines, 
we use three datasets: two small-scale datasets: WebNLG+ 2020 \cite{Ferreira2020The2B} and NYT \cite{zeng2018extracting}, and a large-scale \tekgen dataset from \cite{agarwal2021knowledge}.

\begin{table}
\centering
\caption{WebNLG dataset (Text-to-RDF)}
\label{tab:webnlg}
\begin{tabular}{llll}
                & Train  & Dev   & Test  \\ \hline
RDf triple sets & 13,211 & 1,667 & 752   \\ \hline
Texts           & 35,426 & 4,464 & 2,155 \\
\end{tabular}
\end{table}

\subsection{WebNLG+ 2020}
\label{sec:webnlg_data}
The WebNLG+ corpus v3.0 is part of the 2020 WebNLG Challenge that offers two tasks: the generation of text from a set of RDF triples (subject-predicate-object), and the opposite task of semantic parsing for converting textual descriptions into sets of RDF triples. 
We preprocess the data to remove any underscores and surrounding quotes, in order to reduce noise in the data. Moreover, due to a mismatch of T5 vocabulary and the WebNLG dataset, some characters in WebNLG are not present in T5 vocabulary and ignored during tokenization. We normalize the data mapping the missing characters to the closest available, e.g., `\o' is converted to `o', or `\~a' is mapped to `a'.  




To prepare data for Grapher training, we split the triples into nodes (extracting subjects and objects) and edges (extracting predicates). The nodes are then either sequentially joined as $\langle \textsc{pad} \rangle ~\textsc{node}_1~ \langle \textsc{node\_sep}\rangle ~\textsc{node}_2~ \langle/\textsc{s}\rangle$ for Text Nodes or passed separately as $\langle \textsc{pad} \rangle ~\textsc{node}_1~ \langle/\textsc{s}\rangle$,  $\langle \textsc{pad} \rangle ~\textsc{node}_2~ \langle/\textsc{s}\rangle$ for Query Nodes, padding with $\langle \textsc{no\_node}\rangle$, if necessary. For edges, each element $i, j$ of the adjacency matrix is filled with $\langle \textsc{pad} \rangle ~\textsc{edge}_{i,j}~ \langle/\textsc{s}\rangle$ if there is an edge between $\textsc{node}_i$ and $\textsc{node}_j$ or with $\langle \textsc{pad} \rangle ~\textsc{no\_edge}~ \langle/\textsc{s}\rangle$ otherwise. In case sparse edges are used, we first sparsify the adjacency matrix, and then flatten it to a sequence of edges, similar as for the nodes. Finally, for the classification edge head we scan the training set and collect all the unique predicates to be the edge class list. There are 407 edge classes in our train split, including the $\langle \textsc{no\_edge}\rangle$ class.

\subsection{\tekgen}
\label{sec:tekgen_data}

\tekgen is a large-scale parallel text-graph dataset built by aligning Wikidata KG to Wikipedia text, and its statistics is shown in Table \ref{tab:tekgen_stat}.



\begin{table}
\centering
\caption{Statistics of the \tekgen dataset.}
\label{tab:tekgen_stat}
\begin{tabular}{llll}
         & Train      & Dev     & Test    \\ \hline
Original & 6,383,051  & 797,881 & 797,882 \\ \hline
Processed & 5,391,944 & 673,953 & 678,233 \\
\end{tabular}
\end{table}


\begin{table}
\centering
\caption{Statistics of the NYT dataset.}
\label{tab:nyt_stat}
\begin{tabular}{llll}
         & Train      & Dev     & Test    \\ \hline
Normal & 46,409  & 4,150 & 4,021
\end{tabular}
\end{table}

The data was preprocessed by filtering out triples containing more than 7 predicates, with triple components longer than 100 characters, and with corresponding textual descriptions longer than 200 characters. This was done to match the settings of the WebNLG data and to reduce the computational complexity of the scoring. The final statistics of the dataset is shown in the second row of Table~\ref{tab:tekgen_stat}. In total, the training set contains 1003 predicates/graph edges, which is more than twice larger than in the WebNLG dataset. Note that to match the evaluation to the baseline \cite{dognin2021regen}, and to further manage the limited computational resources, we limit the Test split to 50K sentence-triples pairs.

\subsection{NYT}
\label{sec:nyt_data}
As a third evaluation dataset, we selected the New York Times (NYT) corpus for our experiments, originally proposed by \cite{riedel2010modeling}, consisting of 1.18M sentences. We used an adapted version of the dataset pre-processed by \cite{zeng2018extracting}, referred as "normal", and contains the non-overlapping entities (i.e., head/tail pair has only single edge connecting them), and 25 relation types (the smallest set as compared to WebNLG and \tekgen). Table \ref{tab:nyt_stat} shows the statistics of the dataset.

%% file: experiments.tex
In this Section we provide details about the model setups for evaluations, describe the scoring metrics, and present the results for both datasets.

\subsection{Grapher Setup}
For our base pre-trained language model we used T5 ``large'', for a total number of 770M parameters, from HuggingFace, Inc \cite{wolf2020transformers} (see Appendix for the results using other model sizes). For Query Node generation we also defined the learnable query embedding matrix $M \in \mathbb{R}^{H \times N}$, where $H = 1024$ is the hidden size of T5 model, and $N = 8$ is the maximum possible number of nodes in a graph. The node generation head uses single-layer GRU decoder with $H_{\textsc{GRU}}=1024$ followed by linear transformation projecting to the vocabulary of size $32,128$. The same GRU setup is used for the edge generation head, where we also set the maximum number of edges to be 7. Finally, for the edge classification head, we defined four fully-connected layers with ReLU non-linearities and dropouts with probability 0.5, projecting the output to the space of edge classes.

During training we fine-tuned all the model's parameters, using the AdamW optimizer with learning rate of $10^{-4}$, and default values of $\beta=[0.9, 0.999]$ and weight decay of $10^{-2}$. The batch size was set to 10 samples using a single NVIDIA A100 GPU for WebNLG and NYT training, while for \tekgen training we employed distributed training over 10 A100 GPUs, thus making the effective batch size of 100. Under these settings, it takes approximately 3,500 steps to complete a training epoch for WebNLG, together with the validations done every 1,000 steps, we get a model that reaches its top performance in approximately 6-7 hours. For NYT, the epoch takes approximately 4,600 mini-batches, achieving top performance in about 15 epochs (24 hours). Finally, \tekgen, each epoch takes approximately 54,000 steps, with the evaluations done every 1,000 steps we trained and validated the model for 150,000 iterations, taking approximately 14 days of compute time. 

\subsection{Baselines}

To evaluate the performance of Grapher, for baselines we selected the top performing teams reported on the WebNLG 2020 Challenge Leaderboard, and briefly describe them here:
\textbf{Amazon AI (Shanghai)} \cite{guo-etal-2020-2} was the Challenge winner for Text-to-RDF task. They followed a simple heuristic-based approach that first does entity linking to match the entities present in the input text with the DBpedia ontology, and then query the DBpedia database to extract the relation between them.
\textbf{BT5} \cite{agarwal-etal-2020-machine} came in second place and used large pre-trained T5 model to generate KG in a linearized form, where the object-predicate-subject triples are concatenated together and the entire text-to-graph problem is viewed as a traditional sequence-to-sequence modeling.
\textbf{CycleGT} \cite{Guo2020CycleGTUG}, third place contestant, followed an unsupervised method for text-to-graph and graph-to-text generation, where the KB construction part relies on off-the-shelf entity extractor to identify all the entities present in the input text, and a multi-label classifier to predict the relation between pairs of entities.
\textbf{Stanford CoreNLP Open IE} \cite{manning-etal-2014-stanford}: This is an unsupervised approach that was run on the input text part of the test set to extract the subjects, relations, and objects to produce the output triplets to give a baseline performance for the WebNLG 2020 Challenge.
\textbf{ReGen} \cite{dognin2021regen}: Recent work that leverages T5 pretrained language model and Reinforcement Learning (RL) for bidirectional text-to-graph and graph-to-text generation, which, similarly to  \citet{agarwal-etal-2020-machine}, also follows the linearized graph representation approach.

\begin{table}[t!]
\centering
\caption{Evaluation results on the test set of the WebNLG+ 2020 dataset. The top four block-rows are the results taken from the WebNLG 2020 Challenge Leaderboard \cite{Ferreira2020The2B}. The bottom part shows the results of our proposed Grapher system for several architectural choices, as discussed in Section \ref{sec:method}. Bold font shows the best performing systems.}
\label{tbl:webnlg}
\resizebox{0.48\textwidth}{!}{
\begin{tabular}{ccclccc}
\multicolumn{1}{l}{} & \multicolumn{1}{l}{} & \multicolumn{1}{l}{} & \textbf{M.} & \multicolumn{1}{l}{\textbf{F1}} & \multicolumn{1}{l}{\textbf{Prec.}} & \multicolumn{1}{c}{\textbf{Rec.}} \\ \hline
\multicolumn{3}{l|}{\multirow{3}{*}{\begin{tabular}[c]{@{}c@{}}Amazon AI \end{tabular}}} & \multicolumn{1}{l|}{E} & 0.689 & 0.689 & 0.690 \\
\multicolumn{3}{c|}{}                                                              & \multicolumn{1}{l|}{P}     & 0.696          & 0.696          & 0.698          \\
\multicolumn{3}{c|}{}                                                              & \multicolumn{1}{l|}{S}      & 0.686          & 0.686          & 0.687          \\ \hline
\multicolumn{3}{l|}{\multirow{3}{*}{BT5 }}         & \multicolumn{1}{l|}{E}       & 0.682          & 0.670          & 0.701          \\
\multicolumn{3}{c|}{}                                                              & \multicolumn{1}{l|}{P}     & 0.713          & 0.700          & 0.736          \\
\multicolumn{3}{c|}{}                                                              & \multicolumn{1}{l|}{S}      & 0.675          & 0.663          & 0.695          \\ \hline
\multicolumn{3}{l|}{\multirow{3}{*}{CycleGT }}              & \multicolumn{1}{l|}{E}       & 0.342          & 0.338          & 0.349          \\
\multicolumn{3}{c|}{}                                                              & \multicolumn{1}{l|}{P}     & 0.360          & 0.355          & 0.372          \\
\multicolumn{3}{c|}{}                                                              & \multicolumn{1}{l|}{S}      & 0.309          & 0.306          & 0.315          \\ \hline
\multicolumn{3}{l|}{\multirow{3}{*}{Stanford OIE }} & \multicolumn{1}{l|}{E} & 0.158          & 0.154          & 0.164          \\
\multicolumn{3}{c|}{}                                                              & \multicolumn{1}{l|}{P}     & 0.200          & 0.194          & 0.211          \\
\multicolumn{3}{c|}{}                                                              & \multicolumn{1}{l|}{S}      & 0.127          & 0.125          & 0.130          \\ \hline
\multicolumn{3}{l|}{\multirow{3}{*}{ReGen}}   & \multicolumn{1}{l|}{E}                     & \textbf{0.723} & \textbf{0.714} & \textbf{0.738} \\
\multicolumn{3}{c|}{}                                                              & \multicolumn{1}{l|}{P}     & \textbf{0.767} & \textbf{0.755} & \textbf{0.788} \\
\multicolumn{3}{c|}{}                                                              & \multicolumn{1}{l|}{S}      & \textbf{0.720} & \textbf{0.713} & \textbf{0.735} \\ \hline \\ \hline
\multirow{12}{*}{\begin{tabular}[c]{@{}c@{}}\rotatebox{90}{\textbf{Grapher}}\end{tabular}} &
  \multirow{6}{*}{\begin{tabular}[c]{@{}c@{}}Query\\ Nodes\end{tabular}} &
  \multicolumn{1}{c|}{\multirow{3}{*}{\begin{tabular}[c]{@{}c@{}}Gen\\ Edges\end{tabular}}} & \multicolumn{1}{l|}{E} & 0.395          & 0.391          & 0.400          \\
 &                             & \multicolumn{1}{c|}{}                             & \multicolumn{1}{l|}{P}        & 0.325          & 0.318          & 0.337          \\
 &                             & \multicolumn{1}{c|}{}                             & \multicolumn{1}{l|}{S}         & 0.289          & 0.285          & 0.294          \\ \cline{3-7} 
 &                             & \multicolumn{1}{c|}{\multirow{3}{*}{\begin{tabular}[c]{@{}c@{}}Class\\ Edges\end{tabular}}}
                                                                                   & \multicolumn{1}{l|}{E}          & 0.466          & 0.463          & 0.469          \\
 &                             & \multicolumn{1}{c|}{}                             & \multicolumn{1}{l|}{P}        & 0.360          & 0.356          & 0.368          \\
 &                             & \multicolumn{1}{c|}{}                             & \multicolumn{1}{l|}{S}         & 0.347          & 0.345          & 0.351          \\ \cline{2-7} 
 & \multirow{6}{*}{\begin{tabular}[c]{@{}c@{}}Text\\ Nodes\end{tabular}} & \multicolumn{1}{c|}{\multirow{3}{*}{\begin{tabular}[c]{@{}c@{}}Gen\\ Edges\end{tabular}}}   
                                                                                   & \multicolumn{1}{l|}{E}       & 0.683          & 0.675          & 0.695          \\
 &                             & \multicolumn{1}{l|}{}                             & \multicolumn{1}{l|}{P}     & 0.713          & 0.702          & 0.730          \\
 &                             & \multicolumn{1}{l|}{}                             & \multicolumn{1}{l|}{S}      & 0.681          & 0.673          & 0.693          \\ \cline{3-7} 
 &                             & \multicolumn{1}{c|}{\multirow{3}{*}{\begin{tabular}[c]{@{}c@{}}Class\\ Edges\end{tabular}}} 
                                                                                   & \multicolumn{1}{l|}{E}       & \textbf{0.722} & \textbf{0.715} & \textbf{0.733} \\
 &                             & \multicolumn{1}{c|}{}                             & \multicolumn{1}{l|}{P}     & \textbf{0.750} & \textbf{0.741} & \textbf{0.765} \\
 &                             & \multicolumn{1}{c|}{}                             & \multicolumn{1}{l|}{S}      & \textbf{0.719} & \textbf{0.712} & \textbf{0.730}
\end{tabular}
}
\end{table}

\subsection{Evaluation Metrics}
For scoring the generated graph, we used the evaluation scripts from WebNLG 2020 Challenge \cite{Ferreira2020The2B}, which computes the Precision, Recall, and F1 scores for the output triples against the ground truth. In particular, since the order of generated and ground truth triples should not influence the result, the script searches for the optimal
alignment between each candidate and the reference triple through all possible permutation of the hypothesis-reference pairs. Then, the metrics based on Named Entity Evaluation \cite{SeguraBedmar2013SemEval2013T9} were used to measure the Precision, Recall, and F1 score in four different ways. \textbf{Exact}: The candidate triple should match exactly the reference triple, while the type (subject, predicate, object) is not important. \textbf{Partial}: The candidate triple should match at least partially with the reference
triple, while the type (subject, predicate, object) is irrelevant. \textbf{Strict}: The candidate triple matches exactly the reference triple, and the element type (subject, predicate, object) should match exactly as well. 

\subsection{WebNLG Results}

The main results for evaluating all the compared methods on WebNLG test set are presented in Table~\ref{tbl:webnlg}. As one can see, our Grapher system, based on Text Nodes and Class Edges, achieved on par top performance, as the ReGen \cite{dognin2021regen} model. Our system also uses the Focal loss to account for edge imbalance during training. We can also see that Grapher based on Text Nodes, where the T5-based model generates the nodes directly as a string, outperforms the alternative approach that generates the nodes through query vectors and permutes the features to get invariance to node ordering. A possible explanations is that the graphs at hand and the training data are both quite small. Therefore, the representational power of T5, pre-trained on textual corpora several orders of magnitude larger, can handle the entity extraction task much better. As we mentioned earlier, the ability to extract the nodes is very crucial to the overall success of the system, so if the query-based node generation constructs less reliable sets of nodes, the follow-up stage of edge generation will underperform as well.

\begin{figure}[th!]
\centering
\includegraphics[width=0.48\textwidth]{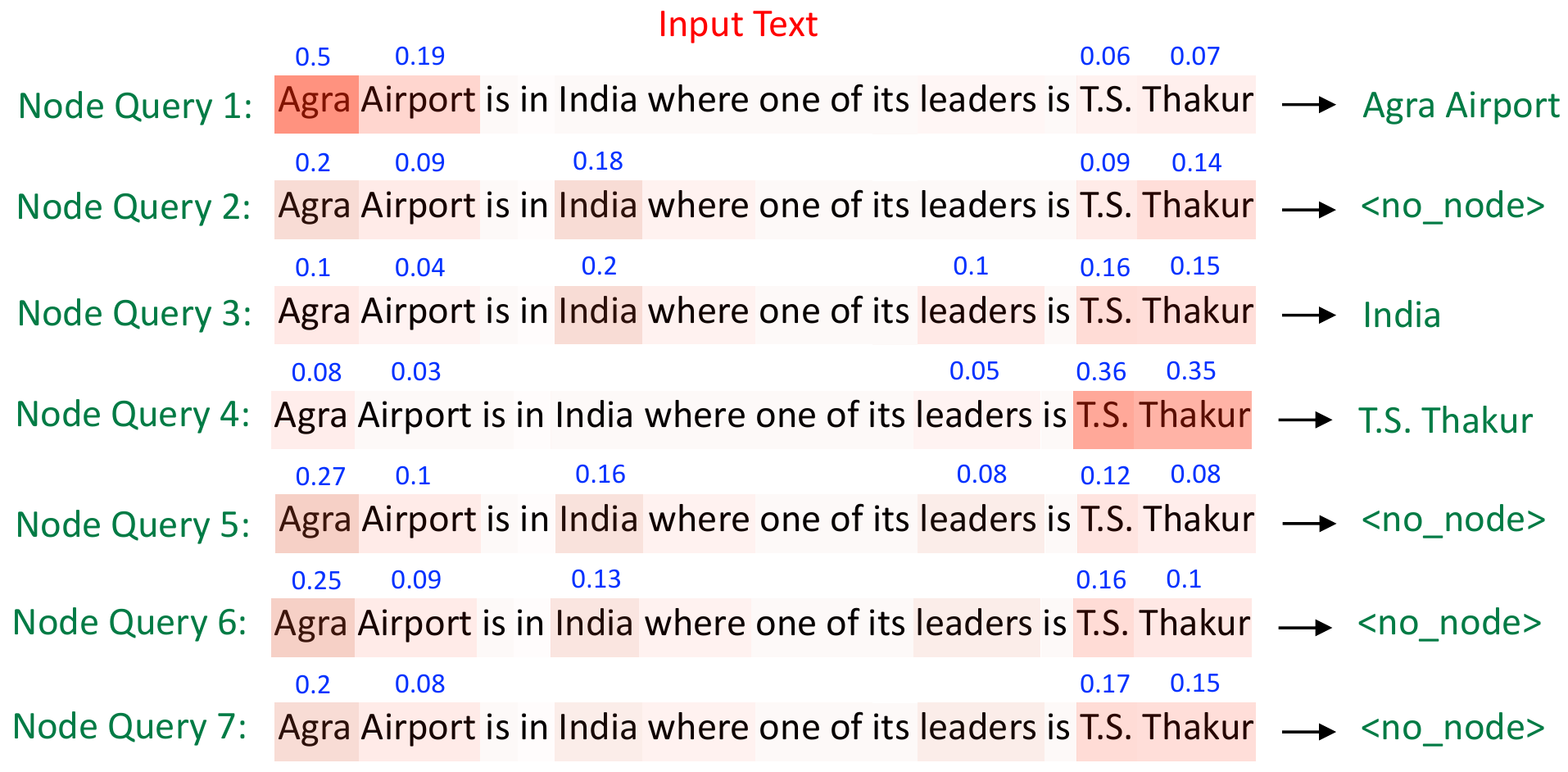}
\caption{Visualization of the cross-attention weights in the T5 model between the node query embedding vectors and the embeddings of the input text.}
\label{fig:node_attention}
\end{figure}


Comparing the edge generation versus classification, we see that the former approach already brings up the system to the level of the top two leaderboard performers, while the edge classification adds extra accuracy and makes Grapher one of the leading system. This again might be due to a smaller training set, in which case GRU edge decoder underperforms, generating less accurate edges, while the classifier just needs to predict a single class to construct an edge, making it a better alternative in the low-data scenarios.    


\begin{table}[t!]
\centering
\caption{Evaluation results on the test set of \tekgen dataset for different configurations of the Grapher system. The use of text-based nodes with generation edges performs the best. }
\label{tab:tekgen}
\bgroup
\def\arraystretch{1.0}%
\resizebox{0.48\textwidth}{!}{
\begin{tabular}{ccccccc}
\multicolumn{1}{l}{} & \multicolumn{1}{l}{} & \multicolumn{1}{l}{} & \textbf{M.} & \multicolumn{1}{l}{\textbf{F1}} & \multicolumn{1}{l}{\textbf{Prec.}} & \multicolumn{1}{c}{\textbf{Rec.}} \\ \hline
\multicolumn{3}{l|}{\multirow{1}{*}{ReGen }}   & \multicolumn{1}{l|}{E}           & 0.623 & 0.610 & 0.647 \\ \hline
\multirow{12}{*}{\begin{tabular}[c]{@{}c@{}}\rotatebox{90}{\textbf{Grapher}}\end{tabular}} &
  \multirow{6}{*}{\begin{tabular}[c]{@{}c@{}}Query\\ Nodes\end{tabular}} &
  \multicolumn{1}{c|}{\multirow{3}{*}{\begin{tabular}[c]{@{}c@{}}Gen\\ Edges\end{tabular}}} & \multicolumn{1}{l|}{E} & 0.386          & 0.361          & 0.430          \\
 &                             & \multicolumn{1}{c|}{}                             & \multicolumn{1}{l|}{P}        & 0.438          & 0.405          & 0.496          \\
 &                             & \multicolumn{1}{c|}{}                             & \multicolumn{1}{l|}{S}         & 0.386          & 0.361          & 0.430          \\ \cline{3-7} 
 &                             & \multicolumn{1}{c|}{\multirow{3}{*}{\begin{tabular}[c]{@{}c@{}}Class\\ Edges\end{tabular}}}
                                                                                   & \multicolumn{1}{l|}{E}          & 0.361          & 0.338          & 0.401          \\
 &                             & \multicolumn{1}{c|}{}                             & \multicolumn{1}{l|}{P}        & 0.408          & 0.378          & 0.463          \\
 &                             & \multicolumn{1}{c|}{}                             & \multicolumn{1}{l|}{S}         & 0.360          & 0.337          & 0.401          \\ \cline{2-7} 
 & \multirow{6}{*}{\begin{tabular}[c]{@{}c@{}}Text\\ Nodes\end{tabular}} & \multicolumn{1}{l|}{\multirow{3}{*}{\begin{tabular}[c]{@{}c@{}}Gen\\ Edges\end{tabular}}}   
                                                                                   & \multicolumn{1}{l|}{E} & \textbf{0.707}   & \textbf{0.693} & \textbf{0.730}\\
 &                             & \multicolumn{1}{l|}{}                             & \multicolumn{1}{l|}{P} & \textbf{0.741}    & \textbf{0.723} & \textbf{0.771} \\
 &                             & \multicolumn{1}{l|}{}                             & \multicolumn{1}{l|}{S} & \textbf{0.706} & \textbf{0.692} & \textbf{0.729}\\\cline{3-7} 
 &                             & \multicolumn{1}{c|}{\multirow{3}{*}{\begin{tabular}[c]{@{}c@{}}Class\\ Edges\end{tabular}}} 
                                                                                   & \multicolumn{1}{l|}{E}      & 0.700 & 0.686 & 0.722 \\
 &                             & \multicolumn{1}{c|}{}                             & \multicolumn{1}{l|}{P}      & 0.735 & 0.717 & 0.764 \\
 &                             & \multicolumn{1}{c|}{}                             & \multicolumn{1}{l|}{S}      & 0.700 & 0.685 & 0.721
\end{tabular}
}
\egroup
\end{table}

Finally, note that although the query-based node generation did not perform well in our evaluations, it is still informative to examine the behaviour of these vectors learned during the training. For this, we analyze the cross-attention weights in the T5 model between the node query vectors and the embeddings of the input text; the results are shown in Fig.~\ref{fig:node_attention}. The ground truth nodes for this sentence are `Agra Airport', `India' and `T.S. Thakur'. It can be seen that each query vector focuses on a set of words that can potentially become a node. For example, the first query vector emphasizes the words `Agra', `Airport', `T.S.' and `Thakur', but since the weight on the first two words is higher, the resulting feature vector sent to the Node GRU module correctly decodes it as `Agra Airport'. The same process happens for the third and forth query vectors. It is also interesting to see that the rest of the queries were also correctly decoded as $\langle \textsc{no\_node}\rangle$ token, even though they had high attention weights on some of the words (e.g., weight of 0.2 on `Agra' and 0.18 on `India' for the second query vector). One potential explanation is that since no causal mask is used when feeding query vectors to the decoder, T5 has an opportunity to exchange the information between all of the query vectors across all the layers and heads. Thus, once the found nodes are assigned to specific vectors, the rest of them are suppressed and decoded into $\langle \textsc{no\_node}\rangle$, irrespective of the attention weights.

\subsubsection{\tekgen Results}

The results on the test set of the \tekgen dataset \cite{agarwal2021knowledge} are shown in Table~\ref{tab:tekgen}. To compute the graph generation performance, we use the same scoring functions as in WebNLG 2020 Challenge \cite{Ferreira2020The2B}. As in Table~\ref{tbl:webnlg}, in this experiment we observe a similar pattern in which the Grapher based on Text Nodes outperforms the query-based system. At the same time we see now that the GRU-based edge decoding performs better than the classification edge head. Recall that for the smaller-size WebNLG dataset the classification edge head performed better, while now on the larger-size \tekgen dataset, the GRU edge generation is more accurate, outperforming the simpler classification edge head. Also, our Grapher model now outperforms the ReGen baseline from \cite{dognin2021regen}, which is based on the linearization technique to represent the graph, showing advantage of the proposed multi-stage generation approach.


\subsubsection{NYT Results}
Finally, Table \ref{tab:nyt} shows the results on NYT dataset. Similar as for the \tekgen, Grapher based on text nodes and generation edges performs the best, outperforming the other architectural choices and the baseline (note that this baseline is our own implementation similar to \cite{dognin2021regen} and \cite{agarwal-etal-2020-machine}, which uses T5 pre-trained language model on the linearized graph representation). Comparing with the results from Tables \ref{tbl:webnlg} and \ref{tab:tekgen}, we can see that for smaller datasets, the classification head has a clear advantage, while as more training data becomes available, the GRU edge decoder becomes more accurate, outperforming the classifier edge head. 

\begin{table}[t!]
\centering
\caption{Evaluation results on the test set of NYT dataset for different configurations of the Grapher system. Text-based nodes with generation edges performs the best.}
\label{tab:nyt}
\bgroup
\def\arraystretch{1.0}%
\resizebox{0.48\textwidth}{!}{
\begin{tabular}{ccccccc}
\multicolumn{1}{l}{} & \multicolumn{1}{l}{} & \multicolumn{1}{l}{} & \textbf{M.} & \multicolumn{1}{l}{\textbf{F1}} & \multicolumn{1}{l}{\textbf{Prec.}} & \multicolumn{1}{l}{\textbf{Rec.}} \\ \hline
\multicolumn{3}{l|}{\multirow{3}{*}{T5 + Linearized Graph}}         & \multicolumn{1}{l|}{E}     & 0.832          & 0.831          & 0.834          \\
\multicolumn{3}{c|}{}                                             & \multicolumn{1}{l|}{P}     & 0.834          & 0.832          & 0.837          \\
\multicolumn{3}{c|}{}                                             & \multicolumn{1}{l|}{S}     & 0.824          & 0.822          & 0.826          \\
\hline
\multirow{6}{*}{\begin{tabular}[c]{@{}c@{}}\rotatebox{90}{\textbf{Grapher}}\end{tabular}}
 & \multirow{6}{*}{\begin{tabular}[c]{@{}c@{}}Text\\ Nodes\end{tabular}} 
                               & \multicolumn{1}{c|}{\multirow{3}{*}{\begin{tabular}[c]{@{}c@{}}Gen\\ Edges\end{tabular}}}  
                                                                            & \multicolumn{1}{l|}{E}     & \textbf{0.918}    & \textbf{0.917} & \textbf{0.920} \\
 &                             & \multicolumn{1}{l|}{}                      & \multicolumn{1}{l|}{P}     & \textbf{0.919}    & \textbf{0.918} & \textbf{0.921}  \\
 &                             & \multicolumn{1}{l|}{}                      & \multicolumn{1}{l|}{S}     & \textbf{0.913}    & \textbf{0.911} & \textbf{0.914}  \\ \cline{3-7} 
 &                             & \multicolumn{1}{c|}{\multirow{3}{*}{\begin{tabular}[c]{@{}c@{}}Class\\ Edges\end{tabular}}} 
                                                                                   & \multicolumn{1}{l|}{E}     & 0.870 & 0.867 & 0.872 \\
 &                             & \multicolumn{1}{c|}{}                             & \multicolumn{1}{l|}{P}     & 0.871 & 0.869 & 0.874 \\
 &                             & \multicolumn{1}{c|}{}                             & \multicolumn{1}{l|}{S}     & 0.860 & 0.858 & 0.862
\end{tabular}
}
\egroup
\end{table}
